\documentclass[10pt,journal,compsoc]{IEEEtran}
\usepackage{amsmath,amsfonts}
\usepackage{algorithmic}
\usepackage{algorithm}
\usepackage{array}
\usepackage[caption=false,font=normalsize,labelfont=sf,textfont=sf]{subfig}
\usepackage{textcomp}
\usepackage{stfloats}
\usepackage{url}
\usepackage{verbatim}
\usepackage{graphicx}
\usepackage{cite}
\usepackage{mathtools}      % \coloneqq

\usepackage{xspace}
\newcommand{\method}{CANOE\xspace}

\newcommand{\traffic}{Traffic Camera\xspace}
\newcommand{\mobile}{Mobile Phone\xspace}

\usepackage{amsopn}

\usepackage{color}

\newcommand{\update}[1]{{\textcolor{black}{#1}}}

\usepackage{enumitem}
\newcommand{\squishlist}
{
	\begin{list}{$\bullet$}
		{
			\setlength{\itemsep}{0pt}
			\setlength{\parsep}{3pt}
			\setlength{\topsep}{3pt}
			\setlength{\partopsep}{0pt}
			\setlength{\leftmargin}{1.5em}
			\setlength{\labelwidth}{1em}
			\setlength{\labelsep}{0.5em}
		}
	}
	
	\newcommand{\squishend}
	{
	\end{list}
}

\usepackage{multirow}
\usepackage{diagbox}
\usepackage{caption}
\usepackage{subcaption}

\usepackage{amssymb}
\usepackage{booktabs}
\usepackage{graphicx}
\usepackage{pifont}

\newsavebox{\measurebox}

\usepackage{multicol}

\usepackage{amsthm}
\usepackage{hyperref}
\newtheoremstyle{named}  
  {0pt}                  
  {0pt}                  
  {\normalfont}          
  {0pt}                  
  {\normalfont} 
  {}                     
  {0pt}    %\newline is for line breaks      
  {}                     

\theoremstyle{named}  
\newtheorem{definition}{Definition}
\newtheorem{problem}{Problem Statement}

\begin{document}

\title{Beyond Regularity: Modeling Chaotic Mobility Patterns for Next Location Prediction}

% CANOE: ChAotic Neural Oscillator nEtwork
% CONCORD: Chaotic Oscillator Network loCatiOn pReDiction

\author{Yuqian~Wu,~
        Yuhong~Peng,~
        Jiapeng~Yu,~
        Xiangyu~Liu,~
        Zeting~Yan,~
        Kang~Lin,~
        Weifeng~Su,~
        Bingqing~Qu,~
        Raymond~Lee,~
        and~Dingqi~Yang% <-this % stops a space
\IEEEcompsocitemizethanks{\IEEEcompsocthanksitem Yuqian Wu, Yuhong Peng, Xiangyu Liu, Zeting Yan, Kang Lin, Weifeng Su, Bingqing QU, and Raymond Lee are with Beijing Normal-Hong Kong Baptist University, Zhuhai, China. Their emails are r130034042@mail.uic.edu.cn, r130034029@mail.uic.edu.cn, t330034034@mail.uic.edu.cn, s230034056@mail.uic.edu.cn, r130034022@mail.uic.edu.cn, wfsu@uic.edu.cn, bingqingqu@uic.edu.cn, and raymondshtlee@uic.edu.cn. \protect\\
Jiapeng Yu is with the University of Warwick, Coventry, United Kingdom. His email is Jiapeng.Yu@warwick.ac.uk. \protect\\
Dingqi Yang is with the University of Macau, Macau, China. His email is 
dingqiyang@um.edu.mo \protect\\
\IEEEcompsocthanksitem Corresponding authors: Bingqing QU (email: bingqingqu@uic.edu.cn) and Raymond Lee (email: raymondshtlee@uic.edu.cn).}
\thanks{Manuscript received April 19, 2021; revised August 16, 2021.}}

% The paper headers
\markboth{Journal of \LaTeX\ Class Files,~Vol.~14, No.~8, August~2021}%
{Shell \MakeLowercase{\textit{et al.}}: A Sample Article Using IEEEtran.cls for IEEE Journals}

\IEEEpubid{0000--0000/00\$00.00~\copyright~2021 IEEE}

\IEEEtitleabstractindextext{%
\begin{abstract}
Next location prediction is a key task in human mobility analysis, crucial for applications like smart city resource allocation and personalized navigation services. 
However, existing methods face two significant challenges: first, they fail to address the dynamic imbalance between periodic and chaotic mobile patterns, leading to inadequate adaptation over sparse trajectories;
second, they underutilize contextual cues, such as temporal regularities in arrival times, which persist even in chaotic patterns and offer stronger predictability than spatial forecasts due to reduced search spaces. 
To tackle these challenges, we propose \textbf{\method}, a 
\underline{\textbf{C}}h\underline{\textbf{A}}otic 
\underline{\textbf{N}}eural \underline{\textbf{O}}scillator n\underline{\textbf{E}}twork for next location prediction, which introduces a biologically inspired Chaotic Neural Oscillatory Attention mechanism to inject adaptive variability into traditional attention, enabling balanced representation of evolving mobility behaviors, and employs a Tri-Pair Interaction Encoder along with a Cross Context Attentive Decoder to fuse multimodal ``who-when-where'' contexts in a joint framework for enhanced prediction performance.
Extensive experiments on two real-world datasets demonstrate that CANOE consistently and significantly outperforms a sizeable collection of state-of-the-art baselines, yielding 3.17\%-13.11\% improvement over the best-performing baselines across different cases. In particular, CANOE can make robust predictions over mobility trajectories of different mobility chaotic levels. A series of ablation studies also supports our key design choices.
Our code is available at: \url{https://github.com/yuqian2003/CANOE}.
\end{abstract}

}

\maketitle

\IEEEdisplaynontitleabstractindextext

\IEEEpeerreviewmaketitle

\IEEEraisesectionheading{\section{Introduction}\label{sec:introduction}}
Next location prediction is a critical yet challenging task in human mobility modeling, serving as a fundamental building block for various location-based services~\cite{wu2025mas4poi,mclp,deepmove,Flashback,graphflash,sgrec,arnn,Wu2024Deep,chen2025selectbalanceplugandplayframework,deng2025revisitingsynthetichumantrajectories,yu2024colearningcodelearningmultiagent}. 
Its core objective is to predict the specific location a user is most likely to visit next, based on the user's historical trajectory data.
Early techniques often resort to traditional stochastic process models, such as Markov Chains and their variants~\cite{mc_app1,mc_app2,mc_app3,fpmc,1mmc}, primarily modeling location transitions through state-dependent probabilistic transition matrices. However, these techniques are inherently limited in capturing long-term and high-order sequential regularities~\cite{mc_setback1,mc_setback2} encoded in user trajectory data. 
Recent techniques mostly resort to deep-learning-based sequence models, with rich expressiveness in capturing the mobility regularities over user trajectories, showing promising results in modeling complex spatiotemporal dependencies.  
For example, they have utilized graph convolutional networks to capture spatial dependencies~\cite{graphflash,sgrec,arnn,stpudgat}, or recurrent neural networks and static attention mechanisms to model temporal and sequential patterns~\cite{getnext,Trans-Aux,mclp,geosan,deepmove,strnn,time_lstm,geosan}. 
\textit{However, these methods primarily focus on capturing the regularity of user mobility behavior by encoding continuous changes in space and time, while overlooking a widely available but underutilized aspect of human movement in predictive systems: chaotic mobility patterns.}

The complex duality of human mobility can be jointly captured by chaotic and periodic patterns~\cite{pappalardo2015returners}. 
This duality manifests itself in two distinct behavior modes: \textit{returners}, whose movements are concentrated around a few key locations such as home or the workplace, embodying periodic patterns; and \textit{explorers}, whose trajectories are scattered across numerous locations, exhibiting chaotic patterns~\cite{pappalardo2015returners, song2010modelling}. 
In other words, while periodic patterns reflect structured routines, such as daily commutes, chaotic patterns capture spontaneous long-range movements, such as travel or social activities. 
Capturing these patterns and their transitions is vital for accurate next location prediction, especially in sparse or long-term datasets where conventional models struggle.
Despite the considerable potential of chaotic mobility pattern recognition, two major challenges persist.

\ding{70} \underline{\textbf{Challenge \uppercase\expandafter{\romannumeral1}: Dynamic Imbalance in Mobility Patterns}}:

\noindent Static attention mechanisms~\cite{geosan,getnext,mobcast,mclp,lda2} assume consistent behavioral patterns and thus fail to capture chaotic mobility patterns while balancing their dynamic interplay with periodic patterns. 
This limitation prevents these mechanisms from adapting to the heterogeneous dynamics of human mobility. 
For periodic patterns, static attention effectively models short-term routines, but cannot adjust when contextual disruptions, such as holidays or social events, alter routine structure or timing, leading to misaligned predictions. 
For chaotic patterns, these mechanisms do not prioritize infrequent long-range visits in sparse trajectories, as temporal discontinuities obscure historical dependencies. 
Crucially, their inherent inflexibility also makes them unable to model transitions between modes, as they cannot dynamically reallocate attention to reflect shifts between chaotic exploration and periodic routines. 

\ding{70} \underline{\textbf{Challenge \uppercase\expandafter{\romannumeral2}: Underutilized Contextual Information}}:
The influence of temporal regularity is critical in modeling human mobility, even within chaotic mobility patterns. 
The next destination of a user can be largely informed by their arrival time. 
For example, an \textit{explorer} might suddenly visit an unusual location in the afternoon due to a hobby. However, if it is evening, she is more likely to go to a nearby supermarket or go home to sleep, rather than to a completely unfamiliar and remote place. 
This type of temporal regularity often exists even in seemingly disordered chaotic trajectories, and its prediction is much more obvious than predicting the next location. This can be attributed to the stronger regularity exhibited in the daily lives of people and a significantly smaller search space in the time dimension compared to the spatial dimension~\cite{huang2015predicting}. 
However, existing methods either fail to explore the usefulness of arrival time, implicitly consider it as an auxiliary task within the next location prediction frameworks, or only attempt to capture temporal patterns of periodic movements~\cite{feng2020predicting,mclp,getnext}, thus overlooking the vast potential of arrival time as direct contextual information. 

To address these limitations, we propose a 
\underline{\textbf{C}}h\underline{\textbf{A}}otic 
\underline{\textbf{N}}eural \underline{\textbf{O}}scillator n\underline{\textbf{E}}twork for next location prediction.
First, to overcome the inflexibility of static attention mechanisms modeling the periodic and chaotic patterns, we introduce a Chaotic Neural Oscillatory Attention  (CNOA), a biologically inspired mechanism that dynamically adjusts attention through excitatory-inhibitory neural interactions, enabling adaptive weighting of routine and exploratory mobility patterns to resolve the dynamic imbalance issue between periodic and chaotic mobile behaviors over sparse trajectory.
Second, to fully leverage contextual information, especially in chaotic mobility scenarios, \method~employs a Tri-Pair Interaction Encoder (TPI-Encoder) that explicitly models pairwise dependencies between multimodal contexts: user-location preferences through LDA~\cite{lda}, user-time alignments via CNOA, and recent location-time dynamics through concatenated embeddings of locations and time slots.
Finally, a Cross Context Attentive Decoder (CCAD) is designed to synthesize these interactions by designating the user-spatial output as the query, while integrating user identity, temporal correlations, and spatial-temporal dynamics as keys and values.
This framework achieves an alignment of ``who-when-where'' dependencies via CNOA, dynamically adapting to individual mobility patterns.
In summary, our main contributions are as follows.
\begin{itemize}[leftmargin=*]
    \item To address the dynamic imbalance issue between chaotic and periodic mobility patterns, we propose a Chaotic Neural Oscillatory Attention mechanism, which uses excitatory-inhibitory neural interactions and an adaptive decay term to dynamically adjust attention weights, learning to balance the contribution of chaotic exploratory patterns and periodic routines for next location prediction.

    \item To fully explore the rich contextual information for location prediction, we design a joint modeling framework that orchestrates the adaptive fusion of multimodal contexts ``who-when-where'' to capture and leverage their synergistic correlations, enabling robust prediction even in chaotic mobility scenarios.
  
    \item We conduct extensive experiments on two real-world datasets to evaluate our CANOE, comparing against a sizeable collection of state-of-the-art methods. 
    Results demonstrate that it significantly outperforms existing solutions, achieving improvements of up to 13.11\% in $Acc@1$ and 10.49\% in $MRR$ on the Traffic Camera dataset, and up to 3.20\% in $Acc@1$ and 3.17\% in $MRR$ on the Mobile Phone dataset, over the best baselines. Our experiments also reveal the robustness of CANOE in making predictions over mobility trajectories of different mobility chaotic levels.
\end{itemize}

\section{Related Work}
\label{sec:related_work}
In this section, we review existing works on next location prediction and chaotic neural oscillator.
\subsection{Next Location Prediction}
\label{sec:relared1}
A plethora of approaches have been proposed to tackle the next location prediction task, broadly categorized into three categories: traditional mathematical methods, sequence-based methods, and graph-based methods. 
Within the realm of traditional mathematical methods, probabilistic frameworks have been extensively utilized to model user mobility, with Markov Chains (MC) and their variants being among the most frequently employed~\cite{mc_app1,mc_app2,mc_app3,fpmc,1mmc}.
These models conceptualize discrete locations as states and construct a transition probability matrix to encode the likelihood of moving between them for individual users.
Early contributions~\cite{mc_early,1mmc} focused on identifying significant locations from GPS data to build Markov models to predict transitions. 
Subsequent advances in Markov models further enhanced prediction performance by incorporating factors such as collective movements~\cite{mc_app2}, location importance~\cite{mc_app3}, and even personalized transition matrices using techniques such as matrix factorization~\cite{fpmc}.
Despite their utility, these traditional techniques inherently face limitations in capturing long-term and high-order sequential regularities due to their core assumption that the current state's dependency is restricted to a limited number of prior steps~\cite{mc_setback1,mc_setback2}.

Graph-based methodologies have been extensively explored for next location prediction. 
Early efforts include STP-UDGAT~\cite{stpudgat}, which employs graph attention networks to learn local and global location correlations.
ARNN~\cite{arnn}, which utilizes meta-paths in knowledge graphs to identify related neighboring locations.
SGRec~\cite{sgrec} combines graph-augmented location sequences with transition patterns between venue categories. 
Prior approaches faced limitations, as they constructed homogeneous graphs with single node and edge types, often neglecting edge weights and focusing solely on connectivity \cite{Flashback}. 
Moreover, when applying GNNs directly on these graphs, the models failed to explicitly learn neighbor importance weights, leading to less interpretable and potentially suboptimal location representations~\cite{lightgcn}. 
More recently, Graph-Flashback~\cite{graphflash} advanced the field by explicitly modeling graph heterogeneity and learning enriched location representations. 
Despite these advancements, Graph-Flashback still lacks user similarity correlations and personalized location transition graphs, limiting its prediction accuracy.

Sequence-based methods have shown promise in modeling spatiotemporal dependencies for next location prediction. Early approaches primarily leveraged recurrent architectures with various enhancements to capture sequential patterns in user trajectories. For instance, ST-RNN~\cite{strnn} integrates spatiotemporal intervals, while Time-LSTM~\cite{time_lstm} augments LSTM with temporal gates. 
To address the challenges posed by sparse and incomplete mobility sequences, several methods incorporate temporal factors by truncating long trajectories into short sessions~\cite{hidasi2015session,feng2018deepmove} or by feeding temporal intervals directly into RNN units~\cite{neil2016phased,zhu2017next}. Moreover, spatiotemporal distances between successive check-ins have been shown to be strong predictors~\cite{yang2014modeling}, leading to models such as HST-LSTM~\cite{kong2018hst} with extended gating mechanisms and STGN~\cite{zhao2020go} with additional gates controlled by spatiotemporal distances.
Flashback~\cite{Flashback} further improves prediction accuracy by aggregating hidden states with similar spatiotemporal contexts, inspiring a series of follow-up works~\cite{cao2021attention, li2021location, wu2022have,liu2022real,rao2022graph} that refine context modeling and hidden state reuse.
Recent advancements have shifted towards attention-based architectures to capture long-range dependencies and complex mobility patterns. GeoSAN~\cite{geosan} enables point-to-point trajectory interactions, GETNext~\cite{getnext} incorporates global trajectory flows with transformers, and MobTcast~\cite{mobcast} utilizes multi-head attention for contextual modeling. However, many of these methods still underutilize explicit spatiotemporal intervals and geographical distances, limiting their ability to capture fine-grained mobility patterns.

Moreover, A significant constraint remains: existing approaches primarily emphasize capturing the periodic mobility routines while overlooking the integration of chaotic behaviors-irregular and exploratory shifts that introduce variability and distant correlations into trajectories, and whose flexible incorporation is crucial for achieving reliable forecasts across diverse and data-scarce scenarios.

\subsection{Chaotic Neural Oscillator}
\label{sec:relared2}
The neural network in the human brain is a complex system comprising approximately 100 billion neurons. Early research focused primarily on small-scale neural networks (2-3 neurons) described by differential or difference equations. 
The solutions to these equations vary over time with changes in initial hyperparameters. 
However, as the network scale increases, solutions fall randomly within certain ranges rather than presenting singular or consistent solutions, leading to chaotic behavior. 

Based on this discovery, the Chaotic Neural Network~\cite{aihara1990chaotic} was first proposed in 1990. Biological experiments have confirmed that neural systems exhibit characteristics such as bifurcation and chaotic attractors. 
Chaotic Neural Networks have become an effective model for processing temporal information by mimicking these properties. Most Chaotic Neural Networks are designed based on classical models such as Hodgkin Huxley~\cite{goldwyn2011what} or Wilson-Cowan~\cite{wilson1972excitatory}; however, these models are either oversimplified to accurately simulate real-world problems or too complex to integrate into artificial neural networks.
Falcke~\cite{falcke2000modeling} subsequently proposed chaotic oscillators, achieving advances in long-term memory, temporal information processing, and dynamic pattern retrieval. 
The Lee oscillator~\cite{Lee2004} was proposed based on spike neural dynamics behavior and excitatory-inhibitory neuron interactions, effectively implementing temporal information processing. 
This work innovatively integrates chaotic neural oscillators into attention mechanisms, proposing Chaotic Neural Oscillatory Attention. 
Unlike the static weight computation of traditional attention mechanisms, we modulate query-key correlations through dynamic excitatory-inhibitory neuron interactions, enabling attention weights to capture both periodic and chaotic mobile patterns in user mobility trajectories.
This biologically inspired dynamic variability allows the model to more accurately capture spatial temporal dependencies, demonstrating particular superiority in modeling user-specific temporal preference patterns.

\section{Preliminary}
\label{secs:statement}
\begin{figure*}[t]
    \centering
    \includegraphics[width=\linewidth]{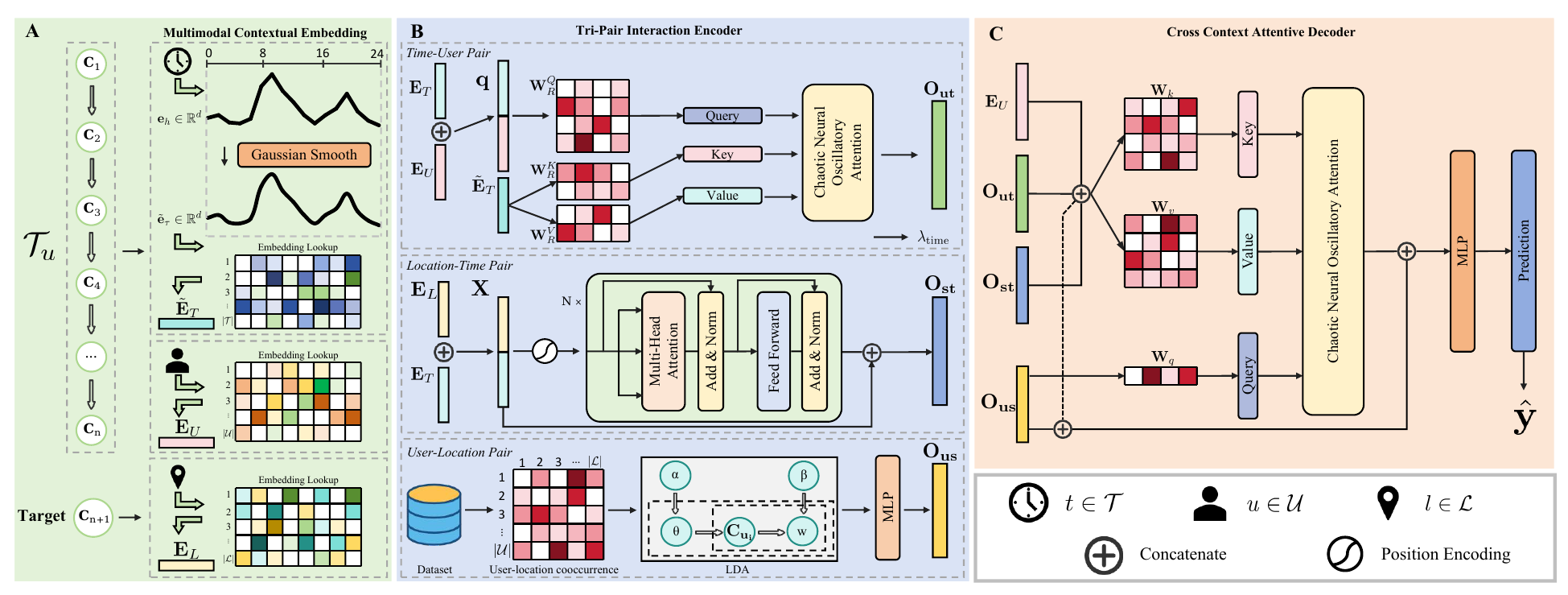}
    \caption{Illustration of the CANOE framework. A) Multimodal Contextual Embedding Module; B) Tri-Pair Interaction Encoder; C) Cross-Context Attentive Decoder, optimized first by the temporal–user pair loss \(\mathcal{L}_{\mathrm{time}}\), then by the location prediction cross-entropy loss \(\mathcal{L}_{\mathrm{loc}}\) and auxiliary classification loss \(\mathcal{L}_{\mathrm{aux}}\) before producing the final next-location prediction.}
    \label{fig:framework}
\end{figure*}

\begin{figure}[t]
    \centering
    \includegraphics[width=\linewidth]{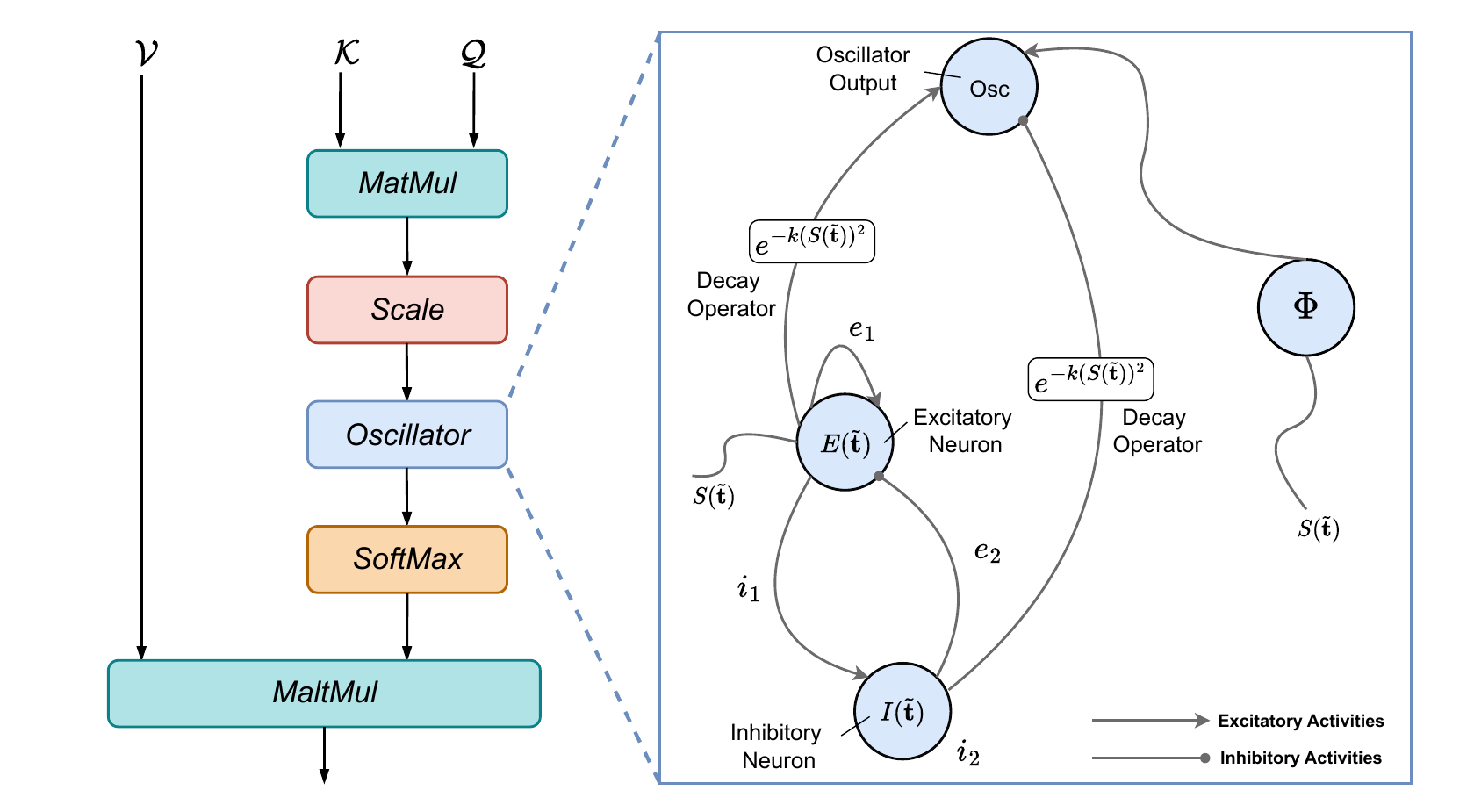}
    \caption{The overview of the Chaotic Neural Oscillatory Attention module.}
    \label{fig:osc}
    \vspace{-3mm}
\end{figure}

Let $\mathcal{U}$, $\mathcal{L}$, and $\mathcal{T}$ denote the sets of users, locations, and timestamps, respectively.
We formalize the next location prediction task as follows.

\begin{definition}[Check-in]
. A \emph{check-in} is defined as a tuple $c = (u, l, t) \in \mathcal{U} \times \mathcal{L} \times \mathcal{T}$, indicating user $u$ visited location $l$ at time $t$;
\end{definition}

\begin{definition}[Trajectory]
. The \emph{trajectory} of user $u$, denoted as $\mathcal{T}_u = \langle c_1, c_2, \dots, c_n \rangle$, is a sequence of check-ins ordered by timestamps in chronological order;
\end{definition}

\begin{definition}[Activity Location] . A location $l \in \mathcal{L}$ is a \emph{activity location} for user $u$ if the continuous dwell duration at $l$ satisfies $\Delta t \geq \theta$ (e.g., $\theta = 60$ minutes);
\end{definition}

\begin{definition}[Sequence]
. The \emph{sequence} $\mathcal{S}^u_{[1:m]} = \langle l_1, l_2, \dots, l_m \rangle$ is derived by extracting all activity locations from $\mathcal{T}^u$, discarding transient visits, here $m$ represents the total number of activity locations;
\end{definition}

\begin{problem}[Next Location Prediction]
. Given a user’s recent subsequence $\mathcal{S}^u_{[n-k:n]} = \langle l_{n-k}, l_{n-k+1}, \dots, l_n \rangle$, within a sliding window of size $k$, infer the subsequent destination $l_{n+1} \in \mathcal{L}$.
\end{problem}

\section{Method}
\label{secs:method}
To address the two key challenges in next location prediction, i.e., dynamic imbalance in mobility patterns and underutilized contextual information, \method~is designed with three main components as shown in Figure~\ref{fig:framework}.

\begin{itemize}[leftmargin=*]
  \item \textbf{Multimodal Contextual Embedding Module} (\S\ref{sec:method_1}).  
  Aiming at transforming heterogeneous ``who-when-where'' contexts into unified vector representations, 
  it generates three distinct embedding matrices $\mathbf{E}_{U}$ for \textit{who}, $\tilde{\mathbf{E}}_{T}$ for \textit{when} and $\mathbf{E}_{L}$ for \textit{where}.

  \item \textbf{Tri-Pair Interaction Encoder} (\S\ref{sec:method_2}).  
  To capture pairwise relationships among user, time, and location modalities, 
  it generates intermediate features $\mathbf{O}{us}$, $\mathbf{O}{ut}$, and $\mathbf{O}_{st}$ that highlight the dynamic interplay between long-term preferences and short-term contextual cues.
  
  \item \textbf{Cross Context Attentive Decoder} (\S\ref{sec:method_3}). 
  It integrates the features of pairwise interaction by aligning ``who-when-where'' dependencies through the CNOA mechanism, generating a robust final representation $\hat{\mathbf{y}}$ for next location prediction.
\end{itemize}

\subsection{Multimodal Contextual Embedding Module}
\label{sec:method_1}
The Multimodal Embedding Module learns joint representations of users, locations, and temporal context to support next location prediction through three dedicated embedding layers.

\noindent\emph{\textbf{Gaussian Smoothed Time Embedding.}} To capture the \emph{periodic regularities} of user behavior encoded in the sparse user trajectory data and address the discontinuity of discrete temporal representations, we discretize one day into $ H = 24 $ time slots, focusing on daily periodicity under an hour granularity. 
Each time slot $h$ is initialized with a learnable embedding vector $\mathbf{e}_h \in \mathbb{R}^{d}$, stacked as $\mathbf{E_T} = [\mathbf{e}_0; \dots; \mathbf{e}_{H-1}] \in \mathbb{R}^{H \times d}$. For a given check-in $c=(u,l,t)$ at slot $\tau \in \{0, \dots, H-1\}$, we compute the \emph{periodic distance} $\Delta(\tau,h) = \min(|\tau-h|, H-|\tau-h|)$ and obtain a smoothed time embedding:
\begin{equation}
\tilde{\mathbf{e}}_{\tau} = \sum_{h=0}^{H-1} w_{\tau,h} \mathbf{e}_h, \quad w_{\tau,h} = \frac{\exp\!\bigl(-\Delta(\tau,h)^2 / 2\sigma^2\bigr)}{\sum_{h'=0}^{H-1} \exp\!\bigl(-\Delta(\tau,h')^2 / 2\sigma^2\bigr)},
\end{equation}
where $\sigma = 1.0$ is empirically set to balance smoothing and specificity. Collecting all $\tilde{\mathbf{e}}_{\tau}$ gives the matrix $\tilde{\mathbf{E}}_{T} = [\tilde{\mathbf{e}}_{0}; \dots; \tilde{\mathbf{e}}_{H-1}] \in \mathbb{R}^{H \times d}$. This smoothing mechanism accounts for the uncertainty of sparse check-in events in a user trajectory by simultaneously encoding not only the time slot of the check-in, but also its cyclical neighboring time slots in the learning process \cite{deng2025replay}.

\noindent\emph{\textbf{User and Location Embedding.}} We employ learnable embedding layers to represent users and locations. The user embedding matrix $\mathbf{E}_U \in \mathbb{R}^{|\mathcal{U}| \times d}$ stores the $d$-dimensional embedding vectors for all users in $\mathcal{U}$, and the location embedding matrix $\mathbf{E}_L \in \mathbb{R}^{|\mathcal{L}| \times d}$ stores the $d$-dimensional embedding vectors for all locations in $\mathcal{L}$:
\begin{align}
\mathbf{E}_U &=
  \bigl[\tilde{\mathbf{e}}_{u_1}; \tilde{\mathbf{e}}_{u_2}; \dots; \tilde{\mathbf{e}}_{u_{|\mathcal{U}|}}\bigr]
  \in \mathbb{R}^{|\mathcal{U}| \times d}, \\[4pt]
\mathbf{E}_L &=
  \bigl[\tilde{\mathbf{e}}_{l_1}; \tilde{\mathbf{e}}_{l_2}; \dots; \tilde{\mathbf{e}}_{l_{|\mathcal{L}|}}\bigr]
  \in \mathbb{R}^{|\mathcal{L}| \times d}.
\end{align}
where $\tilde{\mathbf{e}}_{u_i} \in \mathbb{R}^d$ is the embedding for user $u_i$, and $\tilde{\mathbf{e}}_{l_j} \in \mathbb{R}^d$ is the embedding for location $l_j$.

\vspace{3mm}

\subsection{Tri-Pair Interaction Encoder}
\label{sec:method_2}
We design a Tri‑Pair Interaction Encoder (TPI‑Encoder) that first learns pairwise interactions between each pair of modalities and then aggregates them into a unified representation.

\noindent\emph{\textbf{User-Location Pair.}} 
Capturing the latent patterns in user mobility is crucial for predicting the next location, as it reveals the underlying preferences that guide users' choices. 
However, significant overlap among user trajectory sequences~\cite{sthgcn} poses a major challenge, particularly for \textit{returners}, who frequently visit shared popular locations.
For example, colleagues may exhibit nearly identical trajectories centered around a common workplace and nearby dining venues, resulting in highly similar sequences. 
This inter-user overlap hinders the modeling of unique preferences in user-location pairs, producing overly generalized representations, 
%that struggle to capture individualized patterns, 
especially in the sparse or chaotic trajectories of \textit{explorers} with dispersed movements, thus reducing personalized prediction accuracy.
\cite{sthgcn}
Inspired by~\cite{mclp,lda2}, we address this using LDA~\cite{lda} to model user-level distributions over latent topics, treating users as \textit{documents} and locations as \textit{words}. 
By probabilistically clustering visited locations into latent topics and assigning these topics to users, LDA mitigates this inter-user overlap, enabling distinct preference modeling even in chaotic scenarios.
%It learns to probabilistically cluster the visited locations of user to multiple latent topics and also probabilistically to assign these topics to users for modeling their preference, mitigating the issue of high overlap in location sequences~\cite{sthgcn}. 
%Specifically, let $v_{u_i} = (c_1, c_2, \dots, c_{|\mathcal{L}|})$ denote the histogram of locations visited by user $u_i$, where $c_j$ represents the count of visits to location $l_j \in \mathcal{L}$. The user-location co-occurrence matrix is represented as:
%\begin{equation}
%    C_{u_i} = \text{LDA} \left( [v_{u_1}, \dots, v_{u_{|\mathcal{U}|}}]^T \right)
%\end{equation}
Specifically, let $v_{u_i} = (c_1, c_2, \dots, c_{|\mathcal{L}|})$ denote the histogram of locations visited by user $u_i$, where $c_j$ represents the count of visits to the location $l_j \in \mathcal{L}$. 
We construct the user-location co-occurrence matrix $V \in \mathbb{R}^{|\mathcal{U}| \times |\mathcal{L}|}$ as:
\begin{equation}
    V = [v_{u_1}; v_{u_2}; \dots; v_{u_{|\mathcal{U}|}}]
\end{equation}
LDA is then applied to this matrix to infer the topic distributions:
\begin{equation}
    C_{u_i} = \text{LDA}(V)[i, :]
\end{equation}
Here, $C_{u_i} \in \mathbb{R}^{N^t}$ represents the topic distribution for user $u_i$, where $N^t$ is the predefined number of latent topics.
To further refine the user preference representation, we apply a two-layer MLP to model the interaction between the inferred topics and the user's mobility patterns.
\begin{equation}
\label{eq:us}
    \mathbf{O_{us}}= MLP(C_{u_i})
\end{equation}

\noindent\emph{\textbf{Time-User Pair.}} 
The next destination of a user can be significantly influenced by their arrival time, reflecting temporal regularity even within chaotic mobility patterns.  
However, static attention mechanisms fail to capture the dynamic interplay between periodic and chaotic mobility patterns.
To address this, we propose Chaotic Neural Oscillator Attention (CNOA), a mechanism that introduces adaptive variability to balance the representation of periodic and chaotic patterns by modulating attention weights based on affinity strengths.
Concatenating the user embedding $\mathbf e_{u}$ with the current hour-slot embedding $\mathbf e_{h}$ couples \emph{who} travels with \emph{when}, yielding the query $\mathbf q = [\tilde{\mathbf{e}_{u}} \oplus\,\mathbf e_{h}] \in \mathbb R^{2d}$.
Then Gaussian-smoothed temporal embeddings $\tilde{\mathbf{E}}_{T}$ serve as keys and values, blending adjacent slots and smoothing one-hot jumps.
Projections are computed as:
\begin{equation}
\begin{aligned}
\mathbf{Q}_R = \mathbf{W}_R^Q q, \quad 
\mathbf{K}_R = \mathbf{W}_R^K \tilde{\mathbf{E}}_{T}, \quad 
\mathbf{V}_R = \mathbf{W}_R^V \tilde{\mathbf{E}}_{T} 
\end{aligned}
\end{equation}
where $\mathbf{R}$ denotes the number of attention heads, and $\mathbf{W}_R^Q \in \mathbb{R}^{2d \times d_R}$,  $\mathbf{W}_R^K \in \mathbb{R}^{d \times d_R}$, $ \mathbf{W}_R^V \in \mathbb{R}^{d \times d_R}$ are learnable projection matrices.
The input signal to the oscillator~\cite{Lee2004} is defined as:
\begin{equation}
S(\tilde{\mathbf{t}})=ReLU(\mathbf{Q}_R\mathbf{K}_R^\top)
\end{equation}
where raw affinities are mapping into the oscillator state space.
%internal state space of the oscillator. 
%The dynamics of the excitatory neuron $E(i)$ and the inhibitory neuron $I(i)$ evolve as follows:
Excitatory neurons $E(i)$ and inhibitory neurons $I(i)$ evolve as:
\begin{equation}
\label{eq:osc1}
E(\tilde{\mathbf{t}}+1) = \mathrm{ReLU}\left( e_1 E(\tilde{\mathbf{t}}) + e_2 I(\tilde{\mathbf{t}}) + f(\mathbf{Q}_R\mathbf{K}_R^\top) - \tau_e \right)
\end{equation}
\begin{equation}
I(\tilde{\mathbf{t}}+1) = \mathrm{ReLU}\left( i_1 E(\tilde{\mathbf{t}}) + i_2 I(\tilde{\mathbf{t}}) - \tau_i \right)
\end{equation}
\begin{itemize}[leftmargin=*,noitemsep]
\item $\tilde{\mathbf{t}}\in \{1,2,\dots,\tilde{\mathbf{T}}\}$ indexes the internal oscillator iterations (independent of input sequence length),
\item \(e_1\) is the excitatory self‐feedback strength, \(e_2\) the inhibitory coefficient from \(I\) to \(E\),
\item \(i_1\) is the activation weight from \(E\) to \(I\), \(i_2\) the inhibitory self‐sustain coefficient,
\item \(\tau_e, \tau_i\) are the activation thresholds for \(E\) and \(I\), respectively.
\end{itemize}
For low $S(\tilde{\mathbf{t}})$ (indicative of chaotic patterns with low affinity scores), ReLU activation and feedback loops induce aperiodic fluctuations in $E(\tilde{\mathbf{t}})$ and $I(\tilde{\mathbf{t}})$, increasing diversity in attention weights to capture sparse, exploratory movements.
The oscillator output at step $\tilde{\mathbf{t}}$ is
\begin{equation}
\mathrm{Osc}(\tilde{\mathbf{t}}) = [E(\tilde{\mathbf{t}}+1) - I(\tilde{\mathbf{t}}+1)]e^{-k(S(\tilde{\mathbf{t}}))^2} + \mathrm{ReLU} (S(\tilde{\mathbf{t}}))
\end{equation}
where the decay term $e^{-k S^2(\tilde{\mathbf{t}})}$ (with $k > 0$) modulates the balance: for low $S(\tilde{\mathbf{t}})$, the term approaches 1, amplifying the chaotic difference $[E(\tilde{\mathbf{t}}+1) - I(\tilde{\mathbf{t}}+1)]$ to prioritize diverse long-range dependencies; for high $S(\tilde{\mathbf{t}})$ (indicative of periodic patterns with high affinity scores), the term nears 0, reducing the chaotic contribution and emphasizing the stable $\mathrm{ReLU}(S(\tilde{\mathbf{t}}))$ to focus on recurrent locations. The parameter \(k\) controls the decay rate.%, with a typical value of 500 ensuring smooth transitions.
This differential weighting dynamically adjusts the attention focus based on affinity strength.
The CNOA weights are computed as:
\begin{equation}
\label{eq:osc2}
\alpha_{R,ij}
= \mathrm{Softmax}\!\biggl(\frac{\mathrm{Osc}\bigl(\mathbf{Q}_R^i(\mathbf{K}_R^{j^{^\top}})\bigr)}{\sqrt{d_R}}\biggr)
\end{equation}
with the final output stabilizing temporal attention patterns through:
\begin{equation}
\label{eq:ut}
\mathbf{O_{ut}}
= \sum_{r=1}^{R}
\Bigl[\alpha_R\,\mathbf{V}_R \;\odot\;e^{\bigl(-\gamma\,\lVert \alpha_{R}^{} - \alpha_R^{(\mathrm{prev})}\rVert_2^2\bigr)}\Bigr]
\,\mathbf{W}^O
\end{equation}
%where \(\gamma\) constrains abrupt changes in the attention weights, \(\alpha_R^{(\mathrm{prev})}\) denotes the attention distribution at the previous time step, and
%$\odot\ $ is the elementwise product.
where $\gamma$ constrains abrupt changes, $\alpha_R^{(\mathrm{prev})}$ is the previous attention distribution, and $\odot$ denotes elementwise product. 
The exponential term $e^{-\gamma \|\alpha_R - \alpha_R^{(\mathrm{prev})}\|_2^2}$ smooths transitions by penalizing large deviations from the prior distribution, ensuring gradual shifts between chaotic and periodic patterns as $S(\tilde{\mathbf{t}})$ varies. This adaptive adjustment of attention weights, driven by the oscillator's response to affinity magnitudes, effectively balances the representation of diverse mobility behaviors.

\noindent\emph{\textbf{Location-Time Pair.}} 
Given the recent activity subsequence 
\(\mathcal{S}^u_{[n-k:n]}=\langle l_{n-k},\dots,l_n\rangle\)
with corresponding time slots \(\tau_{n-k},\dots,\tau_n\), we look up
\begin{equation}
\tilde{\mathbf e}_{l_i}=\mathbf E_L[l_i], 
\qquad
\tilde{\mathbf e}_{\tau_i}=\mathbf E_T[\tau_i],
\quad
i=n-k,\dots,n   
\end{equation}
and form  
\begin{equation}
\mathbf x_i=\tilde{\mathbf e}_{l_i} \oplus \tilde{\mathbf e}_{\tau_i},
\qquad
\mathbf X=[\,\mathbf x_{n-k};\dots;\mathbf x_n\,]\in\mathbb R^{(k+1)\times d}
\end{equation}
After applying positional encoding \(\mathrm{PE}(\cdot)\) and a causal mask \(M\), we encode the sequence
\begin{equation}
\mathbf H=\mathrm{Encoder}\bigl(\mathrm{PE}(\mathbf X\sqrt{d}),\,M\bigr)\in\mathbb R^{(k+1)\times d}.
\end{equation}
To preserve both contextualized and fine-grained signals, we concatenate \(\mathbf{O_{st}} = [\,\mathbf H;\,\mathbf X\,]\in\mathbb R^{(k+1)\times 2d}\) as the Location-Time Pair output.

\subsection{Cross Context Attentive Decoder}
\label{sec:method_3}
This module predicts the next location by aligning long-term user preferences with real-time through CNOA mechanism. 
The query vector is derived from the output of the user-location pair $\mathbf{O_{ut}}$, projected as
$\mathbf{O_{us}}\mathbf{W}_q \in \mathbb{R}^d$, to anchor attention on established behavioral patterns.
Keys and values integrate the original user embedding $\tilde{\mathbf{e}}_u$ with time-user pair output $\mathbf{O_{ut}}$ and location-time pair output $\mathbf{O_{st}}$:
\begin{equation}
\begin{aligned}
%\mathbf{Q} &= \mathbf{O_{us}}\mathbf{W}_q \in \mathbb{R}^d\\
\mathbf{K} &= [\tilde{\mathbf{e}}_u \oplus \mathbf{O_{ut}} \oplus \mathbf{O_{st}}]\mathbf{W}_k \in \mathbb{R}^{(d+d+2d) \times d}, \\
\mathbf{V} &= [\tilde{\mathbf{e}}_u \oplus \mathbf{O_{ut}} \oplus \mathbf{O_{st}}]\mathbf{W}_v \in \mathbb{R}^{(d+d+2d) \times d}
\end{aligned}
\end{equation}
where $\mathbf{W}_q \in \mathbb{R}^{d \times d}$, $\mathbf{W}_k \in \mathbb{R}^{(d+d+2d) \times d}$, and $\mathbf{W}_v \in \mathbb{R}^{(d+d+2d) \times d}$ are learnable projection matrices. 
The inclusion of $\tilde{\mathbf{e}}_u$ in keys/values preserves the user's identity characteristics throughout feature transformations, preventing information degradation in deep processing layers. 
The attention weights are calculated by the CNOA mechanism defined in Eq.~\ref{eq:osc1}-~\ref{eq:osc2}:
\begin{equation}
\mathbf{A} = \text{CNOA}(\mathbf{Q}, \mathbf{K}, \mathbf{V})
\end{equation}
The final prediction vector is obtained by applying MLP to the concatenation of multi-source representations:
\begin{equation}
\hat{\mathbf{y}} = \text{MLP}([\mathbf{O_{us}} \oplus \mathbf{O_{st}} \oplus \mathbf{O_{ut}}\oplus \tilde{\mathbf{e}}_u \oplus A])
\end{equation}

\subsection{Next Location Prediction}
Given the final representation $\hat{\mathbf{y}} \in \mathbb{R}^d$ output by the Cross-Context Attentive Decoder, the model computes a probability distribution over the candidate location set $\mathcal{L}$ via a softmax classifier: 
\begin{equation}
P(\hat{l}_{n+1} = l) = \frac{\exp(\mathbf{W}_l^\top \hat{\mathbf{y}} + b_l)}{\sum_{l'\in\mathcal{L}} \exp(\mathbf{W}_{l'}^\top \hat{\mathbf{y}} + b_{l'})}, \quad l \in \mathcal{L}
\label{eq:softmax_loc}
\end{equation}
where $\mathbf{W}_l \in \mathbb{R}^{d}$ and $b_l \in \mathbb{R}$ are learnable parameters. 
The time distribution is estimated using the user-time representation $\mathbf{O_{ut}}$ derived from the time-user pair of the encoder, producing a softmax output over $H = 24$ discretized hourly slots:
\begin{equation}
P(\hat{t}_{n+1} = h) = \frac{\exp(\mathbf{w}_h^\top \mathbf{O_{ut}} + \beta_h)}{\sum_{h'=0}^{H-1} \exp(\mathbf{w}_{h'}^\top \mathbf{O_{ut}} + \beta_{h'})}
\label{eq:softmax_time}
\end{equation}
where $\mathbf{w}_h \in \mathbb{R}^d$ and $\beta_h \in \mathbb{R}$ are learnable projection weights. To reinforce spatial prediction, we introduce an auxiliary classification path that shares the same supervision signal but applies an independent parameter head to the decoder’s representation~\cite{ciampiconi2023survey}. This auxiliary branch provides complementary gradients that can stabilize training and improve top-$k$ retrieval quality.
The final objective combines three components: i) main location prediction, ii) temporal estimation, and iii) auxiliary supervision—weighted by independent hyperparameters:
\begin{equation}
\mathcal{L}_{\text{total}} = \lambda_{\text{loc}} \mathcal{L}_{\text{loc}} + \lambda_{\text{time}} \mathcal{L}_{\text{time}} + \lambda_{\text{aux}} \mathcal{L}_{\text{aux}}
\label{eq:total_loss}
\end{equation}
where:
\begin{align}
\mathcal{L}_{\text{loc}} &= -\frac{1}{|\mathcal{T}|} \sum_{(u,l,t) \in \mathcal{T}} \log P(\hat{l}_{n+1} = l) \\
\mathcal{L}_{\text{time}} &= -\frac{1}{|\mathcal{T}|} \sum_{(u,l,t) \in \mathcal{T}} \log P(\hat{t}_{n+1} = h_g) \\
\mathcal{L}_{\text{aux}} &= -\frac{1}{|\mathcal{T}|} \sum_{(u,l,t) \in \mathcal{T}} \log P_{\text{aux}}(\hat{l}_{n+1} = l)
\end{align}
and $h_g$ is the ground-truth time slot for check-in timestamp $t$. All $\lambda$ weights are hyperparameters tuned to balance the influence of each loss term.

\section{Experiments}
\label{sec:experiments}

\begin{table}[t!]
    \small
    \centering
    \renewcommand{\arraystretch}{0.96}
    \caption{\update{Statistics of experimental datasets.}}
    \label{tab:dataset_stats}
    \resizebox{0.96\columnwidth}{!}{
    \begin{tabular}{c|cccc}
    \toprule
    Dataset & Users & Check-ins & Activity Locations &  Duration \\ \midrule
    \ \traffic~ & 7,800 & 1,115,619 & 2,418 & 61 days \\
    \ \mobile~ & 10,000 & 1,594,551 & 20,607 & 75 days \\
    \bottomrule
    \end{tabular}
    }\vspace{-3mm}
\end{table}

\subsection{Experimental Setup}
\label{sec:Experimental Setup}

\begin{table*}[ht]
\centering
\caption{Performance comparison of \method~  with baseline methods. The results of the baseline models are taken from~\cite{mclp} when applicable. Other results are averaged over five independent runs. The best and runner-up results are highlighted in \textcolor{red}{red} and \textcolor{blue}{blue} respectively. The performance improvement of \method~over the best-performing baselines is also reported.}
\label{tab:overall_performance}
%\small
\resizebox{\textwidth}{!}{
\begin{tabular}{l|cccccccccc}
\toprule
\multirow{2}{*}{Method} & \multicolumn{5}{c}{\traffic~} & \multicolumn{5}{c}{\mobile~} \\
\cmidrule(lr){2-6} \cmidrule(lr){7-11}
 & Acc@1 & Acc@3 & Acc@5 & Acc@10 & MRR & Acc@1 & Acc@3 & Acc@5 & Acc@10 & MRR \\
\midrule
1-MMC & 23.61 & 39.50 & 44.43 & 48.29 & 32.42 & 29.48 & 45.68 & 49.54 & 52.46 & 38.21 \\
Graph-Flashback & 35.69 (0.03) & 55.64 (0.08) & 63.73 (0.05) & 72.25 (0.05) & 48.18 (0.03) & 37.61 (0.03) & 59.62 (0.02) & 65.86 (0.02) & 71.88 (0.03) & 50.31 (0.02) \\
SNPM & 36.43 (0.05) & 56.18 (0.02) & 63.74 (0.04) & 71.45 (0.02) & 48.58 (0.01) & 37.99 (0.07) & 59.89 (0.02) & 66.03 (0.04) & 71.92 (0.01) & 50.60 (0.04) \\
DeepMove & 35.89 (0.07) & 51.60 (0.07) & 57.72 (0.07) & 65.15 (0.04) & 46.08 (0.05) & 37.38 (0.04) & 56.84 (0.03) & 63.10 (0.02) & 69.88 (0.03) & 49.11 (0.02) \\
Flashback & 34.89 (0.06) & 54.92 (0.09) & 62.88 (0.04) & 71.00 (0.07) & 47.33 (0.04) & 37.39 (0.03) & 59.64 (0.05) & 65.96 (0.04) & 72.01 (0.06) & 50.22 (0.02) \\
STAN & 29.92 (0.10) & 49.70 (0.12) & 57.81 (0.08) & 66.24 (0.10) & 42.39 (0.08) & 36.40 (0.09) & 56.43 (0.10) & 62.15 (0.12) & 67.77 (0.12) & 48.06 (0.09) \\
GETNext & 35.53 (0.11) & 54.26 (0.14) & 61.27 (0.21) & 68.64 (0.12) & 47.15 (0.11) & 36.93 (0.09) & 59.44 (0.11) & 65.75 (0.40) & 71.80 (0.64) & 49.89 (0.09) \\
Trans-Aux & 36.69 (0.12) & 53.97 (0.13) & 60.38 (0.11) & 67.50 (0.07) & 47.52 (0.09) & 38.52 (0.31) & 56.66 (0.11) & 61.76 (0.18) & 67.28 (0.18) & 49.24 (0.12) \\
CSLSL & 36.96 (0.12) & 55.02 (0.13) & 61.67 (0.11) & 68.79 (0.07) & 48.17 (0.09) & 37.86 (0.16) & 60.22 (0.07) & 66.52 (0.02) & 71.94 (0.01) & 50.51 (0.11) \\
MCLP-LSTM & 39.90 (0.06) & 58.32 (0.07) & 65.14 (0.07) & 72.43 (0.07) & 51.28 (0.05) & 39.42 (0.16) & 60.74 (0.07) & 66.95 (0.06) & 72.98 (0.06) & 51.81 (0.08) \\
MCLP-Attention & \textcolor{blue}{40.11 (0.05)} & \textcolor{blue}{58.44 (0.05)} & \textcolor{blue}{65.30 (0.04)} & \textcolor{blue}{72.58 (0.05)} & \textcolor{blue}{51.46 (0.02)} & \textcolor{blue}{39.65 (0.02)} & \textcolor{blue}{61.02 (0.05)} & \textcolor{blue}{67.18 (0.06)} & \textcolor{blue}{73.15 (0.05)} & \textcolor{blue}{52.04 (0.03)} \\

\midrule
\method~ & \textcolor{red}{45.37 (0.09)} & \textcolor{red}{64.43 (0.18)} & \textcolor{red}{71.03 (0.15)} & \textcolor{red}{77.78 (0.16)} & \textcolor{red}{56.86 (0.12)} & \textcolor{red}{40.92 (0.01)} & \textcolor{red}{63.04 (0.12)} & \textcolor{red}{69.41 (0.07)} & \textcolor{red}{75.49 (0.07)} & \textcolor{red}{53.69 (0.04)} \\

Improvement (\%) & 13.11 & 10.25 & 8.78 & 7.16 & 10.49 & 3.20 & 3.30 & 3.32 & 3.20 & 3.17 \\
\bottomrule
\end{tabular}
}
\end{table*}

\subsubsection{Datasets.}
We evaluate \method~ on two real-world datasets, namely \traffic~ (\textbf{TC}) \cite{traffic} and \mobile~(\textbf{MP}) \cite{mobile}, in the next location prediction tasks.
The \traffic~dataset comprises vehicle trajectory data reconstructed from over 3,000 traffic cameras, capturing spatiotemporal movement patterns of approximately 5 million vehicles during four days in 2021–2022. 
The \mobile~ dataset aggregates anonymized human mobility trajectories of 100,000 individuals over 90 days, discretized into 500m × 500m grid cells (200 × 200 grid in total) to ensure privacy, with timestamps masked into 30-minute intervals. 
In data preprocessing, we followed~\cite{mclp,stan,Flashback} by excluding sequences with fewer than 100 location records and segmenting the remaining sequences into fixed-length subsequences of 20 timesteps. 
These subsequences are then partitioned into training, validation, and testing sets at a ratio of 7:1:2 per user. 
We define activity locations as grid cells or intersections where a user remains for at least one continuous hour. 
Note that due to the protection of user privacy, the exact GPS coordinates in the two datasets are not accessible. 
Table \ref{tab:dataset_stats} summarizes the key statistics of the preprocessed datasets.

\subsubsection{Baselines.}
We consider the following baselines from three categories. 
1) \textit{Markov Chain-based model.} \textbf{1-MMC}~\cite{1mmc}: A first-order Mobility Markov Chain defines the current location as the state, with transition probabilities representing the likelihood of moving between locations.
2) \textit{Graph-based models.} \textbf{Graph-Flashback}~\cite{graphflash}: It uses Graph Convolutional Networks on a location transition graph to encode POI representations, which are subsequently utilized in RNN-based models for predicting user locations. \textbf{SNPM}~\cite{snpm}: The Sequence-based Neighbour Search and Prediction Model extracts POI relationships from check-in data using knowledge graph embedding and Eigenmap methods, and builds a dynamic neighbor graph to aggregate similar POIs, leveraging previous states for predictions.
3) \textit{Sequential-based models.} \textbf{DeepMove}~\cite{deepmove}: An attentional recurrent neural network model for predicting human mobility from lengthy and sparse trajectories, capturing sequential transition patterns and multi-level periodicity through a multi-modal embedding and a historical attention mechanism. \textbf{Flashback}~\cite{Flashback}: An RNN architecture designed to model sparse user mobility traces by referencing past hidden states with similar spatiotemporal contexts. \textbf{STAN}~\cite{stan}: It leverages a bi-layer attention mechanism to capture spatiotemporal correlations across user trajectories and personalize predictions based on item frequency. \textbf{GETNext}~\cite{getnext}: It utilizes a user-agnostic global trajectory flow map for effective POI embedding generation. \textbf{Trans-Aux}~\cite{Trans-Aux}: Trans-Aux is a transformer decoder-based neural network that predicts an individual's next location by considering historical locations, time, and travel modes, with the prediction of travel mode as an auxiliary task to guide the learning process. \textbf{CSLSL}~\cite{cslsl}: It utilizes causal multi-task learning to model the "time→activity→location" decision logic and a spatial-constrained loss function to align predicted and actual destination distributions, while capturing transition regularities across different time spans. \textbf{MCLP-LSTM}~\cite{mclp}: It explicitly incorporates user preferences and arrival time as context, utilizing a topic model, an arrival time estimator, and a LSTM architecture to integrate multiple contextual information and mine sequential patterns. \textbf{MCLP-Attention}~\cite{mclp}: A variant of MCLP-LSTM that replaces the LSTM network with an Transformer encoder for extracting sequential information.

\subsubsection{Experimental Metrics}
We utilize the standard evaluation metrics for location prediction tasks, in line with previous studies~\cite{mclp,getnext}: Top-k accuracy (Acc@k, \( k \in \{1, 3, 5, 10\} \)) and Mean Reciprocal Rank (MRR). 
Given a test dataset with \( N_{test} \) samples, the evaluation metrics are defined as:
\[
\begin{aligned}
    \text{Acc@k} &= \frac{1}{N_{test}} \sum_{i=1}^{N_{test}} 1(r_i < k) \quad &\text{MRR} &= \frac{1}{N_{test}} \sum_{i=1}^{N_{test}} \frac{1}{rank_i}
\end{aligned}
\]
Acc@k measures the proportion of correct predictions within the top \( k \) rankings, while MRR reflects the overall ranking quality, with higher values indicating better performance.

\subsubsection{Implement Details}
Models are trained for 100 epochs with a batch size of 256 and an embedding dimension of 16 for the TC dataset and 8 for the MP dataset, using PyTorch 2.1.2 on NVIDIA A100 GPUs under Ubuntu 22.04. Optimization is carried out using AdamW with a learning rate of 0.005 and a weight decay of 0.01. 
We employed grid search to find the optimal weights for $\mathcal{L}_{\text{loc}}$, $\mathcal{L}_{\text{time}}$, and $\mathcal{L}_{\text{aux}}$
within the range of [0, 1].
In the location–time pair, we stack three layers of the Transformer encoder with a dropout rate of 0.1. 
The number of latent topics is set to 450 for the TC dataset and 400 for the MP dataset. 
In addition, for the CNOA mechanism, we apply a specific parameter configuration: we set $k=-500$ throughout all experiments. The remaining parameters, including the excitatory self-feedback strength $e_1$, the inhibitory coefficient $e_2$, the activation weight $i_1$, the inhibitory self-sustain coefficient $i_2$, and the activation thresholds $\tau_e$ and $\tau_i$, are determined by grid search within predefined ranges. Specifically, $e_1, e_2 \in \{-1, 0, 1\}$, while $i_1, i_2$, and $\tau_e, \tau_i$  are searched in [-10, 10].
We also conduct sensitivity analysis on key hyperparameters in Section~\ref{sec:exp_parameter} and present the results for \method~ and the best-performing baseline models in Table~\ref{tab:overall_performance}.

\begin{table*}[t!]
    \centering
    \renewcommand{\arraystretch}{0.96}
    \caption{\update{Performance comparison over different chaotic levels}.} %\traffic}.}
    \label{tab:chaotic}
    \small
    \begin{tabular}{c|c|l|ccccc}
    \toprule
    \textbf{Dataset} & 
    \textbf{Threshold} & \textbf{Model} & \textbf{Acc@1} &\textbf{Acc@3} & \textbf{Acc@5} & \textbf{Acc@10} & \textbf{MRR} \\ \midrule
    \multirow{16}{*}{\traffic} &
    \multirow{4}{*}{0.75}
& MCLP-LSTM & 39.50 (0.06) & 57.78 (0.16) & 64.55 (0.01) & 71.75 (0.02) & 50.80 (0.01)\\
& & MCLP-Attention & 39.67 (0.06) & 57.93 (0.06) & 64.73 (0.02) & 71.90 (0.06) & 50.97 (0.02) \\
& & \method(w/o CNOA) & 43.80 (0.05) & 63.12 (0.24) & 70.07 (0.34) & 77.27 (0.31) & 55.58 (0.12) \\
& & \method & 45.00 (0.11) & 64.10 (0.16) & 70.77 (0.26) & 77.58 (0.27) & 56.54 (0.14) \\ 
\cmidrule{2-8} %
& \multirow{4}{*}{0.80}
& MCLP-LSTM & 38.78 (0.07) & 57.19 (0.16) & 64.06 (0.01) & 71.36 (0.02) & 50.18 (0.01) \\
& & MCLP-Attention & 38.97 (0.05) & 57.34 (0.08) & 64.24 (0.02) & 71.51 (0.06) & 50.36 (0.02) \\
& & \method(w/o CNOA) & 43.12 (0.05) & 62.61 (0.24) & 69.67 (0.35) & 76.97 (0.31) & 55.02 (0.12) \\
& & \method & 44.38 (0.10) & 63.63 (0.16) & 70.40 (0.27) & 77.30 (0.27) & 56.03 (0.14) \\
\cmidrule{2-8} %

& \multirow{4}{*}{0.85} & MCLP-LSTM & 37.27 (0.07) & 55.84 (0.15) & 62.89 (0.01) & 70.39 (0.01) & 48.82 (0.01) \\
& & MCLP-Attention & 37.49 (0.04) & 56.02 (0.06) & 63.07 (0.02) & 70.56 (0.05) & 49.03 (0.01) \\
& & \method(w/o CNOA) & 41.72 (0.07) & 61.43 (0.26) & 68.69 (0.36) & 76.23 (0.32) & 53.81 (0.14) \\
& & \method & 43.09 (0.11) & 62.55 (0.17) & 69.51 (0.29) & 76.64 (0.28) & 54.92 (0.15) \\
\cmidrule{2-8} %

& \multirow{4}{*}{0.90} & MCLP-LSTM & 35.40 (0.12) & 53.96 (0.16) & 61.13 (0.03) & 68.89 (0.00) & 47.03 (0.03) \\
& & MCLP-Attention & 35.69 (0.03) & 54.15 (0.06) & 61.38 (0.02) & 69.10 (0.05) & 47.29 (0.00) \\
& & \method(w/o CNOA) & 40.04 (0.06) & 59.79 (0.27) & 67.22 (0.37) & 75.03 (0.36) & 52.25 (0.16) \\
& & \method & 41.63 (0.13) & 61.17 (0.17) & 68.27 (0.31) & 75.65 (0.32) & 53.57 (0.17) \\
    \bottomrule
%%%%%%%%%%%%%%%%%%%%%%%%%%%%%
    \multirow{16}{*}{\mobile} &
    \multirow{4}{*}{0.75}
& MCLP-LSTM & 37.01 (0.05) & 59.08 (0.15) & 65.31 (0.10) & 71.41 (0.05) & 49.77 (0.07) \\
& & MCLP-Attention & 37.34 (0.01) & 59.33 (0.08) & 65.70 (0.15) & 71.92 (0.40) & 50.09 (0.03) \\
& & CANOE(w/o CNOA) & 37.82 (0.06) & 60.85 (0.00) & 67.63 (0.02) & 74.16 (0.03) & 51.16 (0.04) \\
& & CANOE & 38.76 (0.04) & 61.45 (0.08) & 68.01 (0.09) & 74.34 (0.08) & 51.87 (0.05) \\
\cmidrule{2-8}

& \multirow{4}{*}{0.80}
& MCLP-LSTM & 36.20 (0.05) & 58.39 (0.15) & 64.70 (0.11) & 70.87 (0.05) & 49.04 (0.07) \\
& & MCLP-Attention & 36.56 (0.00) & 58.66 (0.08) & 65.09 (0.16) & 71.38 (0.41) & 49.39 (0.04) \\
& & CANOE(w/o CNOA) & 37.05 (0.06) & 60.18 (0.01) & 67.07 (0.02) & 73.69 (0.03) & 50.48 (0.05) \\
& & CANOE & 38.02 (0.08) & 60.83 (0.06) & 67.49 (0.10) & 73.90 (0.05) & 51.21 (0.05) \\
\cmidrule{2-8}

& \multirow{4}{*}{0.85}
& MCLP-LSTM & 35.17 (0.06) & 57.41 (0.15) & 63.82 (0.11) & 70.09 (0.04) & 48.07 (0.09) \\
& & MCLP-Attention & 35.56 (0.04) & 57.71 (0.10) & 64.25 (0.17) & 70.60 (0.44) & 48.44 (0.06) \\
& & CANOE(w/o CNOA) & 36.07 (0.10) & 59.22 (0.01) & 66.24 (0.03) & 73.00 (0.03) & 49.55 (0.06) \\
& & CANOE & 37.06 (0.09) & 59.88 (0.04) & 66.69 (0.07) & 73.24 (0.04) & 50.32 (0.05) \\
\cmidrule{2-8}

& \multirow{4}{*}{0.90}
& MCLP-LSTM & 34.32 (0.05) & 56.27 (0.18) & 62.70 (0.12) & 69.05 (0.05) & 47.11 (0.08) \\
& & MCLP-Attention & 34.78 (0.08) & 56.62 (0.09) & 63.15 (0.18) & 69.57 (0.44) & 47.54 (0.09) \\
& & CANOE(w/o CNOA) & 35.26 (0.12) & 58.10 (0.08) & 65.22 (0.01) & 72.11 (0.00) & 48.66 (0.09) \\
& & CANOE & 36.25 (0.05) & 58.82 (0.10) & 65.72 (0.09) & 72.36 (0.07) & 49.43 (0.05) \\
    \toprule
    
    \end{tabular}
    % }
    \vspace{-3mm}
\end{table*}

\subsection{Overall Performance}
\label{sec:exp_overall} 
Table~\ref{tab:overall_performance} shows the performance comparison of our model against baselines for the next location prediction. We see that CANOE consistently and significantly outperforms all baselines, in particular, yielding 3.17\%-13.11\% improvement over best-performing baselines across different cases. We discuss detailed results below.
\ding{182} Markov Chains variants (1-MMC) consistently exhibited the worst performance, achieving $Acc@1$ scores of only 23.61\% on the \traffic~ and 29.48\% on the \mobile.
This aligns with established benchmarks for next location prediction and can be attributed to their inherent limitations. Specifically, 1-MMC primarily models location transitions using state-dependent probabilistic transition matrices, restricting their capacity to capture long-term and higher-order sequential patterns embedded in user trajectory data.
\ding{183} While methods like DeepMove and Trans-Aux utilize attention mechanisms to capture mobility patterns and demonstrate relative improvements, with $Acc@1$ and $MRR$ ranging from 35.89\% to 36.69\% and 46.08\% to 47.52\%, respectively, in \traffic, and from 37.38\% to 38.52\% and 49.11\% to 49.24\% in \mobile, they still struggle to effectively balance dynamic shifts between periodic and chaotic patterns.
\ding{184} Although MCLP-LSTM and MCLP-Attention achieved the best performance among the baselines, with $Acc@1$ and $MRR$ ranging from 39.90\% to 40.11\% and 51.28\% to 51.46\%, respectively, in \traffic, and from 39.42\% to 39.65\% and 51.81\% to 52.04\% in \mobile, their effectiveness remains constrained by their handling of temporal context. 
These methods primarily emphasize capturing periodic temporal patterns, such as daily or weekly routines, which represent only a subset of the intricate dynamics inherent in human mobility. 
For instance, while an individual might predictably return home in the evening, they could also visit a novel location during the day, a seemingly chaotic event in which the time of day nonetheless offers essential predictive context. 
Moreover, these models fail to integrate chaotic information. 
They process distinct information streams, including location sequences and temporal features, as isolated components, and the absence of a unified aggregation mechanism hinders their ability to fully exploit the rich and nuanced temporal context embedded in trajectory data.
\ding{185} By effectively capturing and balancing periodic and chaotic mobile patterns and aligning the ``who-when-where'' dependencies, our \method~achieves the best performance across all metrics in the two datasets. 
\begin{table*}[t]
\centering
\caption{Ablation study on \traffic~and \mobile. The Tri-Pair Interaction (TPI) Encoder column specifies the activated branches using abbreviations: UL (User–Location), TU (Time–User), and LT (Location–Time). 
A \ding{51} in the Time Emb column indicates the use of Gaussian Smoothed Temporal Embedding; otherwise, a 24-hour lookup embedding is applied. 
A \ding{51} in the Chaotic Neural Oscillatory Attention (CNOA) column denotes the use of CNOA, while \ding{55} corresponds to cross attention. 
A \ding{51} in the Decoder column indicates the use of the Cross-Context Attentive Decoder (CCAD); otherwise, a two-layer MLP is used. 
Results are averaged over five independent runs.}
\label{tab:ablation}
\small
% \resizebox{1.0\textwidth}{!}{
\begin{tabular}{c|cccc|ccccc}
\toprule
\textbf{Dataset} & \textbf{Time Emb} & \textbf{CNOA} & \textbf{TPI Encoder} & \textbf{CCAD} & \textbf{Acc@1} &\textbf{Acc@3} & \textbf{Acc@5} & \textbf{Acc@10} & \textbf{MRR} \\ \midrule

\multirow{6}{*}{TC}       
& \ding{55} & \ding{55} & LT & \ding{55} & 29.48 (0.16) & 42.65 (0.18) & 48.39 (0.15) & 55.71 (0.17) & 38.53 (0.16) \\
& \ding{55} & \ding{55} & LT, UL & \ding{55} & 38.07 (0.04) & 54.96 (0.10) & 61.28 (0.07) & 68.38 (0.10) & 48.62 (0.16) \\
& \ding{55} & \ding{55} & LT, UL, TU & \ding{55} & 40.24 (0.09) & 58.35 (0.03) & 65.21 (0.08) & 72.55 (0.14) & 51.51 (0.09) \\
& \ding{55} & \ding{55} & LT, UL, TU & \ding{51} & 44.19 (0.27) & 63.69 (0.24) & 70.61 (0.24) & 77.61 (0.23) & 55.99 (0.24)\\
& \ding{51} & \ding{55} & LT, UL, TU & \ding{51} & 44.53 (0.09) & 63.87 (0.08) & 70.73 (0.07) & 77.58 (0.11) & 56.25 (0.09) \\
& \ding{51} & \ding{51} & LT, UL, TU & \ding{51} & 45.37 (0.09) & 64.43 (0.18) & 71.03 (0.15) & 77.78 (0.16) & 56.86 (0.12) \\ \midrule

\multirow{6}{*}{MP}    
& \ding{55} & \ding{55} & LT & \ding{55} & 25.55 (0.49) & 41.00 (0.38) & 46.95 (0.25) & 54.51 (0.47) & 35.49 (0.07) \\
& \ding{55} & \ding{55} & LT, UL & \ding{55} & 38.87 (0.07) & 60.19 (0.02) & 66.39 (0.02) & 72.38 (0.06) & 51.25 (0.06) \\
& \ding{55} & \ding{55} & LT, UL, TU & \ding{55} & 39.43 (0.01) & 60.43 (0.02) & 66.63 (0.02) & 72.61 (0.01) & 51.67 (0.01) \\
& \ding{55} & \ding{55} & LT, UL, TU & \ding{51} & 40.09 (0.02) & 62.41 (0.03) & 68.97 (0.01) & 75.24 (0.03) & 53.02 (0.03) \\
& \ding{51} & \ding{55} & LT, UL, TU & \ding{51} & 40.19 (0.07) & 62.58 (0.05) & 69.09 (0.05) & 75.28 (0.03) & 53.12 (0.06) \\
& \ding{51} & \ding{51} & LT, UL, TU & \ding{51} & 40.92 (0.01) & 63.04 (0.12) & 69.41 (0.07) & 75.49 (0.07) & 53.69 (0.04) \\ \bottomrule
\end{tabular}
% }
\end{table*}

\subsection{Performance over Different Chaotic Levels}
In this experiment, we evaluate the capability of \method~in capturing chaotic mobility patterns. We first present our settings below, followed by the results. Specifically, at each time step $\tau$ in a test trajectory, we can compute the entropy of locations
$H_{\tau} = -\sum_{\ell \in \mathcal{L}^{\mathrm{pref}}_{\tau}} p_{\ell}(\tau)\,\ln p_{\ell}(\tau)$
using only the empirical distribution of location on the distinct prefix set $\mathcal{L}^{\mathrm{pref}}_{\tau}$, where
$p_{\ell}(\tau) = f_{\ell}(\tau-1)/(\tau-1)$ denotes the empirical probability of location $\ell$ in the prefix (with $f_{\ell}(\tau-1)$ denoting its frequency).
This value is normalized as $\hat{H}_{\tau} = H_{\tau} / \ln m_{\tau} \in [0,1]$, where $m_{\tau} = |\mathcal{L}^{\mathrm{pref}}_{\tau}|$ denotes the number of distinct locations observed up to step $\tau$.
Subsequently, this entropy measures the extent to which the chaotic (high-entropy) or periodic (low-entropy) mobility patterns are relevant in moving to the next location. We then segment the trajectories using a series of entropy thresholds $\Theta = \{0.75, 0.80, 0.85, 0.90\}$ to distinguish different levels of chaotic intensities, and compare the performance of \method~against the top two best-performing baselines (MCLP-LSTM and MCLP-Attention) and an ablated variant of our \method using cross attention instead of CNOA (denoted as \method(w/o CNOA)), across different levels.

Table~\ref{tab:chaotic} shows the results across different entropy thresholds. We observe that: \ding{182} \method~consistently outperformed baseline models in all evaluation metrics, demonstrating its ability to handle chaotic movement patterns. 
\ding{183} As movement patterns became increasingly chaotic (with thresholds increasing from 0.75 to 0.90), \method~exhibited significantly smaller performance degradation compared to all baseline models. 
For instance, \method~ decreased by 3.37\% in $Acc@1$, while MCLP-LSTM, MCLP-Attention and \method(w/o CNOA) decreased by 4.10\%, 3.98\%, and 3.76\%, respectively. This result implies that our \method~ with the CNOA module can effectively learn to balance the chaotic or periodic mobility patterns for predicting the next location, thus better adapting to cases of different levels of chaotic intensities.
This trend is consistent in all evaluation metrics, highlighting the robust resilience of \method~ in chaotic mobility scenarios.
\subsection{Ablation Study}
In this section, we conduct a systematic ablation study evaluating the utility of our key design choices, including 1) Gaussian Smoothed Time Embedding (\textbf{Time Emb}), 2) Chaotic Neural Oscillatory
Attention (\textbf{CNOA}), 3) \textbf{TPI Encoder}, and 4) Cross Context Attentive Decoder (\textbf{CCAD}).

Table~\ref{tab:ablation} shows the results. We discuss the observations and their implications below. \ding{182} Compared to the 24-hour lookup embedding, the Gaussian Smoothed Temporal Embedding achieves improvements, with Acc@1 and MRR ranging from 44.19\% to 44.53\% and 55.99\% to 56.25\%, respectively, on \traffic, and from 40.09\% to 40.19\% and 53.02\% to 53.12\% on \mobile.
The Gaussian Smoothed Temporal Embedding enhances modeling of transitions near time slot boundaries (e.g., midday or late evening check-ins) by integrating contributions from adjacent slots through a weighted representation. 
This smoothing mechanism encodes both the check-in time slot and its cyclical neighbors, mitigating the uncertainty of sparse check-in events and the representation fragmentation inherent in 24-hour lookup embeddings.
\ding{183} The integration of the CNOA mechanism yields substantial performance improvements, with $Acc@1$ and $MRR$ ranging from 44.53\% to 45.37\% and 56.25\% to 56.86\%, respectively, on \traffic, and from 40.19\% to 40.92\% and 53.12\% to 53.69\% on \mobile.
This enhancement mitigates the challenge of dynamic imbalance in mobility patterns by enabling the model to robustly capture long-range dependencies and transitions between periodic and chaotic behaviors.
\ding{184} Progressive activation of the three branches (LT, UL, TU) of the TPI Encoder leads to consistent improvements across all metrics in \traffic and \mobile. 
The addition of the UL branch, which leverages user historical mobility data, allows the model to move beyond generalized spatiotemporal patterns and more precisely capture individual-level preferences and unique transition tendencies. 
Furthermore, the TU branch introduces a crucial temporal dimension by modeling the probabilistic estimation of the next arrival time of a user.
This empowers the model to refine its predictions by harnessing time as a direct contextual signal, which is often more predictable than location itself. 
\ding{185} By introducing the Cross Context Attentive Decoder (CCAD) to adaptively integrate the contributions of the three encoder branches, our model achieves $Acc@1$ improvements of 9.81\% and 8.70\% on \traffic~ and \mobile, respectively, compared to baseline methods.
The User-Location (UL) branch captures stable, long-term user preferences, reflecting habitual mobility patterns. In contrast, the Location-Time (LT) and Time-User (TU) branches provide dynamic contextual signals, encoding immediate ``where-when'' and ``who-when'' relationships for each check-in.
CCAD effectively balances these stable and fluid signals, leveraging chaotic attention to dynamically model their complex interplay, thereby enhancing next-location prediction accuracy. 

\subsection{Impact of Different Attention Mechanisms}
\label{sec:exp_strategy}
\begin{table*}[t!]
    \centering
    \renewcommand{\arraystretch}{0.96}
    \caption{\update{Different attention mechanism performance comparison on \traffic.}}
    \label{tab:strategy}
    \small
    % \resizebox{0.96\columnwidth}{!}{
    \begin{tabular}{c|ccccc}
    \toprule
    \textbf{Strategy} & \textbf{Acc@1} &\textbf{Acc@3} & \textbf{Acc@5} & \textbf{Acc@10} & \textbf{MRR} \\ \midrule
    \ Self Attention & 36.47 (0.50) & 54.46 (0.18) & 61.25 (0.11) & 68.50 (0.15) & 47.70 (0.34) \\
    \ Cross Attention & 44.53 (0.09) & 63.87 (0.08) & 70.73 (0.07) & 77.58 (0.11) & 56.25 (0.09) \\
    \ Linear & 29.44 (0.04) & 58.32 (0.06) & 66.95 (0.02) & 74.84 (0.00) & 56.25 (0.01) \\
    \ STAN & 41.17 (0.31) & 63.65 (0.34) & 72.27 (0.26) & 80.92 (0.16) & 54.84 (0.28) \\
    \ CNOA & 45.37 (0.09) & 64.43 (0.18) & 71.03 (0.15) & 77.78 (0.16) & 56.86 (0.12) \\
    %\ \mobile~ & 10,000 & 1,594,551 & 20,607 & 75 days \\
    \bottomrule
    \end{tabular}
    % }
    % \vspace{-4mm}
\end{table*}

In this experiment, we study the impact of different attention mechanisms to showcase the advantages of CNOA over static attention for next location prediction. 
Specifically, to ensure a fair comparison, we replace only the attention component, while keeping all other architectural and training configurations unchanged. We consider the following alternative attention mechanisms: 1) Self Attention~\cite{vaswani2017attention}, 2) Cross Attention~\cite{huang2019ccnet}, 3) Linear Attention~\cite{wang2020linformer}, and 4) STAN's Attention~\cite{stan}. It should be noted that, due to the fact that our datasets encode locations as anonymized IDs (without geographic coordinates), we implement a STAN attention that realizes only STAN~\cite{stan}’s trajectory-aggregation layer and injects a scalable Personalized Item Frequency (PIF) bias, omitting the explicit spatial distance used in STAN’s bi-layer design.

Table~\ref{tab:strategy} shows the results. We observe that CNOA outperforms other competing attention mechanisms in all evaluation metrics, especially in $Acc@1$ (45.37\%) and $MRR$ (56.86\%). Specifically, Although linear attention reduces computational complexity via kernel-based softmax approximation, it performs worst in $Acc@1$ (29.44\%). STAN attention achieves the best Acc@5 (72.27\%) and Acc@10 (80.92\%) by aggregating spatiotemporal correlations within trajectories through a dual-layer self-attention architecture and injecting scalable PIF biases to emphasize revisit probabilities. Cross-attention leverages temporally aligned contextual interactions through query-key separation. However, these mechanisms remain statically dependent, thus unable to fully address the non-stationary transitions inherent in chaotic motion. In contrast, CNOA embeds the dynamic properties of neural oscillators into the attention weight generation process. Through an excitation-inhibition cycle, chaotic fluctuations are introduced to low-affinity signals while maintaining periodic attention to high-affinity signals, thereby achieving adaptive balancing for heterogeneous motion patterns. This highlights the critical importance of adaptive weight allocation for distinguishing sparse, chaotic trajectories from dense, periodic trajectories.

\begin{figure}[]
    \centering
    \includegraphics[width=0.50\textwidth]{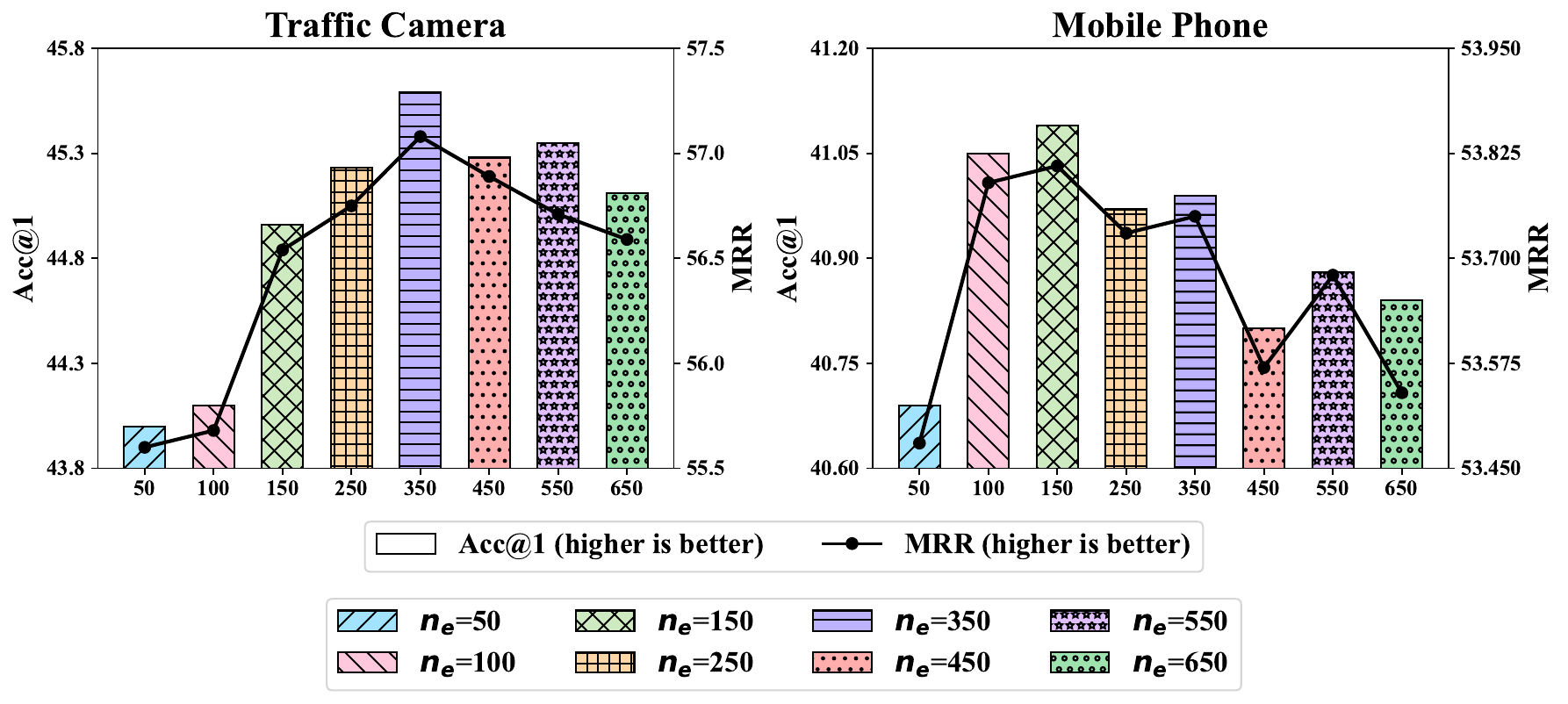}
    \caption{Impact of topic numbers on the prediction performance.}
    \label{fig:tn_exp}
    %\vspace{-2.5em}
\end{figure}

\subsection{Parameter Study}
\label{sec:exp_parameter}
In this section, we conduct a parameter study to evaluate the impact of three key hyperparameters on the performance of \method.

\subsubsection{Impact of the number of latent topics.}  
As Figure~\ref{fig:tn_exp} shows, we observe that when the topic number is low, the number of latent topics may be insufficient to capture the diversity of user behaviors, potentially leading to the compression of multiple distinct movement patterns into a few topics, thus hurting the prediction performance. 
In contrast, when the topic number is excessively high, it generates redundant latent topics, each potentially representing an over-specific user behavior pattern or location combinations, and thus reduces the generalization ability, with this effect being particularly pronounced in the MP dataset, leading to a clear drop in performance as the topic number increases. 
In \traffic, a topic number of 350 may strike a balance, with Acc@1 and MRR reaching 45.59\% and 57.08\%, respectively, effectively leveraging the advantages of fine-grained topics while avoiding the drawbacks of excessive dispersion.%, though performance slightly decreases when the topic number increases to 450, 550, and 650. 
In \mobile, a lower topic number of 150 is sufficient to capture key patterns, with Acc@1 and MRR at 41.09\% and 53.81\%, respectively, while an excess of topics may introduce noise, adversely affecting prediction performance.
%\vspace{-10mm}
\begin{figure}[t]
    \centering
\includegraphics[height=0.60\linewidth]{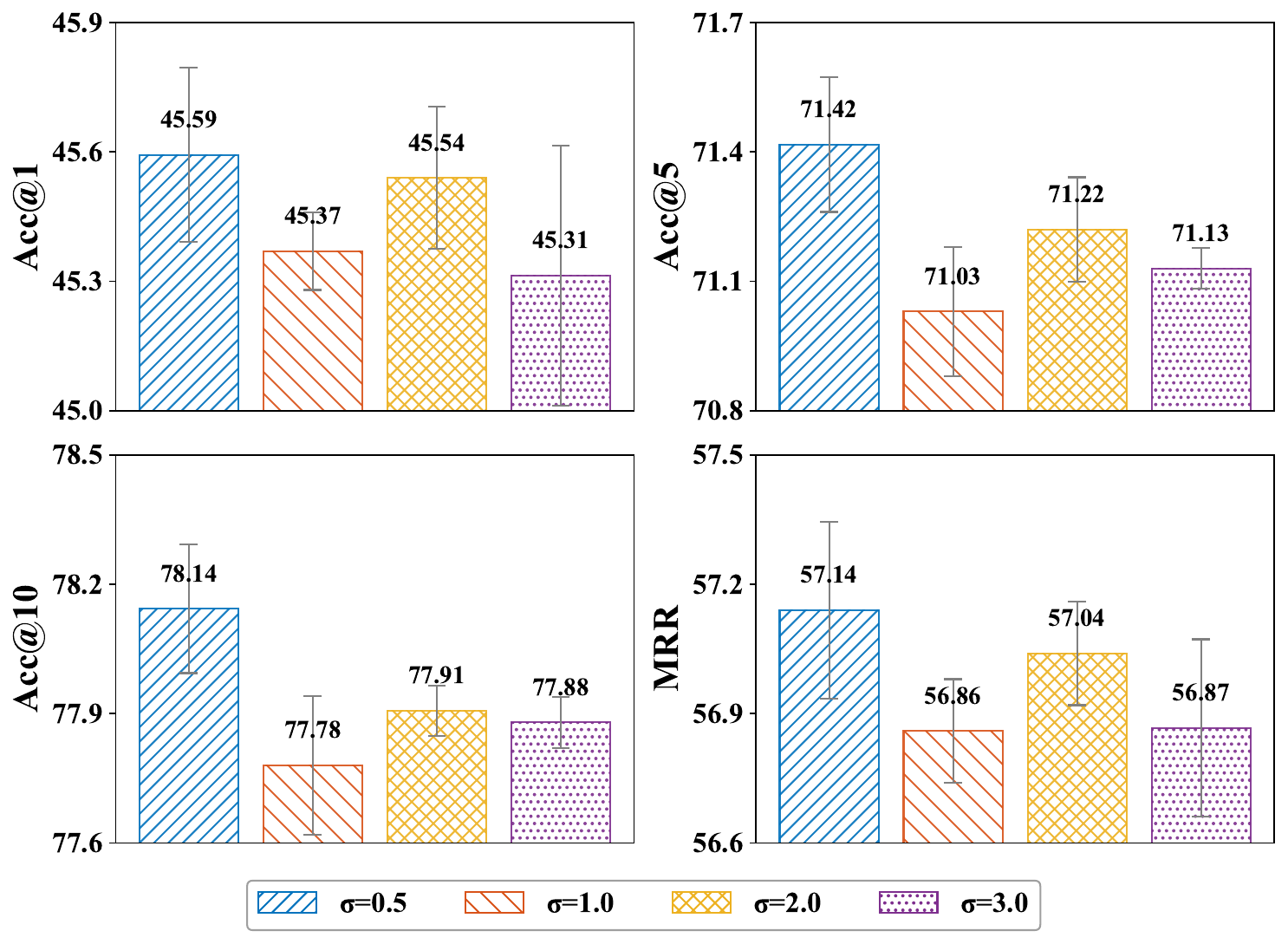}
    \caption{Different $\sigma $ value performance comparison on \traffic~.}
    \label{fig:sigma}
    \vspace{-5mm}
\end{figure}
%\vspace{-8mm}

\subsubsection{Impact of the $\sigma$ value for the Gaussian smoothed time embedding.}
As Figure~\ref{fig:sigma} shows, we observe that the configuration with $\sigma = 0.5$ consistently achieves the highest performance across all evaluation metrics, yielding $Acc@1$ of 45.59\%  and $MRR$ of 57.14\%. When $\sigma$ is small (e.g., 0.5), the Gaussian kernel becomes sharp, concentrating weights on the current time slot $\tau$ and a few adjacent slots, with contributions to distant slots rapidly decaying to near zero. 
This smoothing strategy mitigates the hard boundaries of discrete representations (e.g., 24-hour lookup embedding) while avoiding excessive blurring. 
In sparse check-in data, user behavior often exhibits strong daily periodicity and specific hourly preferences. 
A $\sigma=0.5$ preserves high-resolution temporal cues, ensuring that embeddings capture precise periodic patterns without blending unrelated time slots (e.g., conflating morning and afternoon behaviors). 
In contrast, larger $\sigma$ values expand the smoothing radius, homogenizing embeddings and introducing noise (e.g., leaking non-peak behaviors into peak periods), which reduces the model sensitivity to behavioral transitions (e.g., midday patterns) and degrades downstream prediction accuracy.
\begin{figure}[]
    \centering
\includegraphics[height=0.60\linewidth]{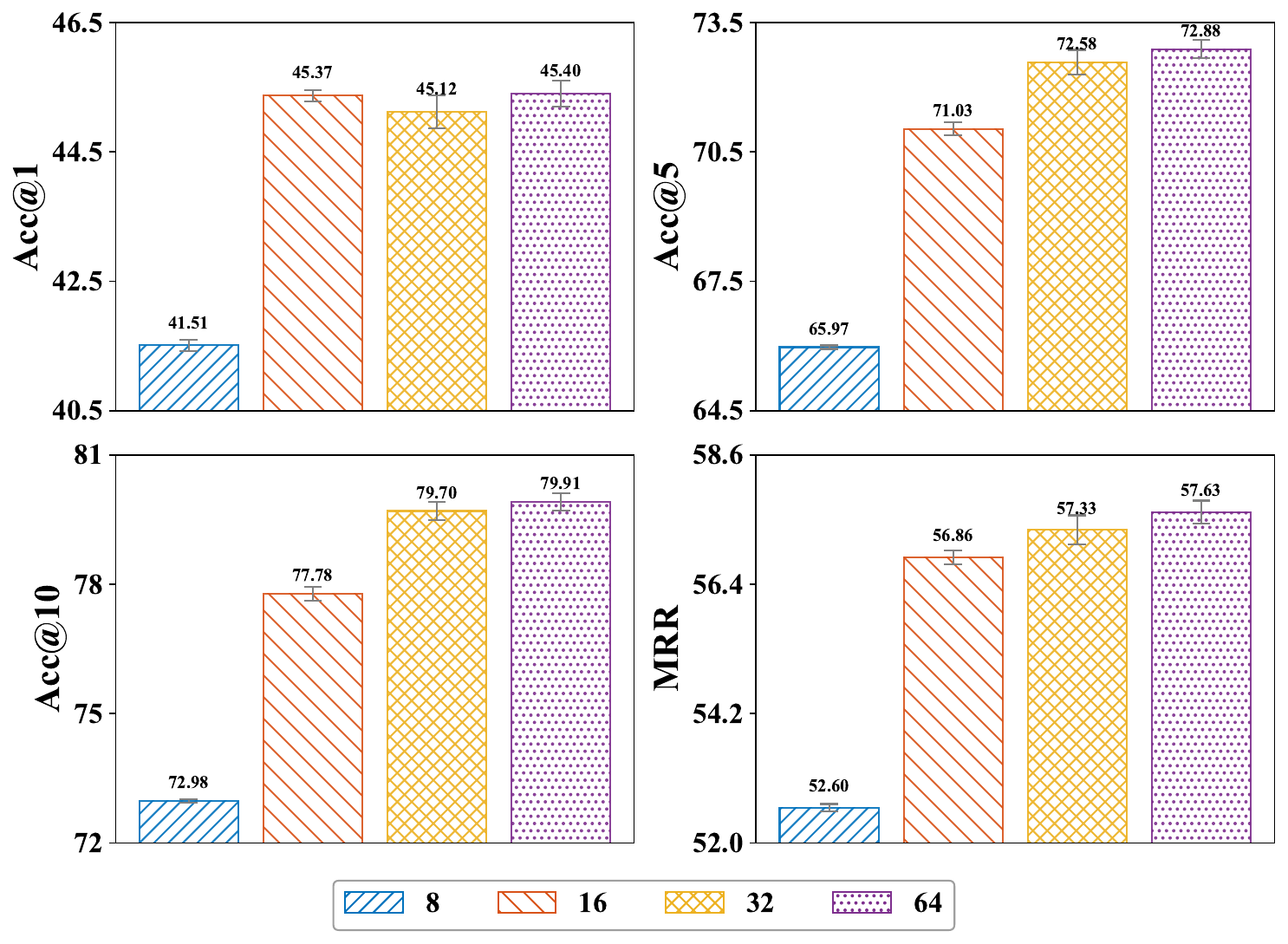}
    \caption{Different embedding size performance comparison on \traffic~.}
    \label{fig:emb}
\end{figure}

\subsubsection{Impact of the embedding size.} As shown in Figure~\ref{fig:emb}, we see that the embedding dimension \(d=64\) achieves optimal performance with Acc@1 (45.40\%) and MRR (57.63\%), while \(d=8\) shows a significant drop due to low-dimensional embeddings not capturing the mobile pattern, resulting in information loss. The close performance of \(d=32\) and \(d=64\) suggests diminishing returns from higher dimensions.

\section{Conclusion}
\label{sec:conclusion}
This paper addresses the challenges of balancing chaotic and periodic mobile patterns and investigating the underutilized contextual information for next location prediction. To this end, we introduce \method~, a 
\underline{\textbf{C}}h\underline{\textbf{A}}otic 
\underline{\textbf{N}}eural \underline{\textbf{O}}scillator n\underline{\textbf{E}}twork for next location prediction, which integrates a biologically inspired Chaotic Neural Oscillatory Attention mechanism to inject adaptive variability into traditional attention, enabling balanced representation of evolving mobility behaviors, and employs a Tri-Pair Interaction Encoder along with a Cross Context Attentive Decoder to fuse multimodal ``who-when-where'' contexts in a joint framework for enhanced prediction performance. We conduct extensive experiments against a sizeable collection of state-of-the-art baselines on two real-world datasets. Results demonstrate that CANOE consistently and significantly outperforms state-of-the-art baselines, yielding 3.17\%-13.11\% improvement over best-performing baselines across different cases. In particular, we also see that CANOE can make robust predictions over mobility trajectories of different mobility chaotic levels by adaptively learning to balance the chaotic and periodic mobility patterns when predicting the next location. A series of systematic ablation studies also supports our key design choices.

In the future, we plan to further investigate the impact of balancing the two types of mobility patterns in a broader range of spatiotemporal prediction tasks.

\section{Acknowledgement}
The authors thank for Beijing Normal-Hong Kong Baptist University and the IRADS lab for the provision of computing facilities for the conduct of this research. This project has received funding from the Science and Technology Development Fund, Macau SAR (0011/2025/RIB1, 001/2024/SKL), and Jiangyin Hi-tech Industrial Development Zone under the Taihu Innovation Scheme (EF2025-00003-SKL-IOTSC).

%\section{Reference}
%\balance
\bibliographystyle{IEEEtran}
\bibliography{references.bib}

\vfill

\end{document}